\documentclass[11pt, pre, aps, floatfix, onecolumn]{revtex4-2}%

\usepackage[left=1in, right=1in, top=0.75in, bottom=1in, footskip=0.25in]{geometry}
\usepackage{amsmath}
\usepackage{amssymb}
\usepackage{setspace}
\usepackage{hyperref}

\begin{document}

\title{Improved Dimensionality Reduction for Inverse Problems in Nuclear Fusion and High-Energy Astrophysics}
\author{Jonathan Gorard}
\email{gorard@princeton.edu}
\affiliation{Princeton Plasma Physics Laboratory, Princeton, NJ, USA}
\author{Ammar Hakim}
\email{ahakim@pppl.gov}
\affiliation{Princeton Plasma Physics Laboratory, Princeton, NJ, USA}
\author{Hong Qin}
\email{hongqin@princeton.edu}
\affiliation{Princeton Plasma Physics Laboratory, Princeton, NJ, USA}
\author{Kyle Parfrey}
\email{kparfrey@pppl.gov}
\affiliation{Princeton Plasma Physics Laboratory, Princeton, NJ, USA}
\author{Shantenu Jha}
\email{shantenu@pppl.gov}
\affiliation{Princeton Plasma Physics Laboratory, Princeton, NJ, USA}

\maketitle

\vspace{-0.9cm}

\noindent
\textbf{Topic: Optimization algorithms for inverse problems under uncertainty}

\vspace{-0.6cm}

\section{Executive Summary}

\vspace{-0.5cm}

Many inverse problems in nuclear fusion and high-energy astrophysics research, such as the optimization of tokamak reactor geometries or the inference of black hole parameters from interferometric images, necessitate high-dimensional parameter scans and large ensembles of simulations to be performed. Such inverse problems typically involve large uncertainties, both in the measurement parameters being inverted and in the underlying physics models themselves. Monte Carlo sampling, when combined with modern non-linear dimensionality reduction techniques such as autoencoders and manifold learning, can be used to reduce the size of the parameter spaces considerably. However, there is no guarantee that the resulting combinations of parameters will be physically valid, or even mathematically consistent. In this position paper, we advocate adopting a hybrid approach that leverages our recent advances in the development of formal verification methods for numerical algorithms, with the goal of constructing parameter space restrictions with provable mathematical and physical correctness properties, whilst nevertheless respecting both experimental uncertainties and uncertainties in the underlying physical processes.

\vspace{-0.75cm}

\section{Key Challenges}

\vspace{-0.5cm}

Many fundamental tasks in nuclear fusion and high-energy astrophysics research take the form of inverse problems, necessitating large-scale, high-dimensional parameter scans. When designing magnetic confinement fusion reactors, such as tokamaks and stellarators, one is typically attempting to optimize for certain key plasma and reactor performance parameters, such as energy confinement time and plasma beta, with respect to many aspects of the reactor's design, such as its elongation, aspect ratio, or other features of its geometry\cite{freidberg_designing_2015}. The large numbers of possible magnetic equilibrium configurations in tokamaks, coil geometries in stellarators, etc. naturally give rise to very large and high-dimensional parameter spaces of possible reactor designs. Similarly large parameter space searches arise in the inference of black hole parameters, such as spin and inclination angle, from interferometric images of black hole shadows\cite{broderick_themis:_2020}; or in the inference of neutron star equation of state parameters, such as baryonic number density and pressure, from macroscopic astrophysical observables, such as total mass, radius, and tidal deformability\cite{ferreira_extracting_2022}. Moreover, due to the high temperatures, intense neutron fluences, and radiation fluxes inside a fusion device, and the impacts of the intervening medium upon Earth-bound measurements of astrophysical phenomena, the measurements being inverted in each of these cases are typically sparse, noisy, and associated with large experimental uncertainties.

Naively, solving such inverse problems necessitates sampling these parameter spaces in a uniform fashion, typically by running large ensembles of numerical simulations. However, in each of the aforementioned cases, the high dimensionality of the parameter space often renders such a task computationally intractable unless the simulations are run at prohibitively low resolution (thus compromising physical fidelity), or the dimensionality of the parameter space can be substantially reduced. Fortunately, many classical dimensionality reduction techniques, such as principal component analysis and linear discriminant analysis in the linear case, or (variational) autoencoders and manifold learning in the non-linear case, may be used to project the high-dimensional parameter space down to a lower-dimensional subspace that nevertheless preserves many of its salient features\cite{sorzano_2014}. In this way, such dimensionality reduction methods can implicitly identify reduced collections or combinations of simulation parameters with the highest degree of physical impact or relevance, whilst potentially only requiring a small Monte Carlo sample of high-resolution, high-fidelity simulations to be run. Such Monte Carlo samples can also be effectively biased so as to reflect experimental noise or other uncertainties in the measurement data, as well as any inherent uncertainties in the physics models themselves.

However, an important limitation of classical dimensionality reduction techniques is that they remain agnostic to the details of the problem being solved, and hence there is no guarantee that the lower-dimensional subspace onto which one projects will be physically (or even mathematically) valid. For example, there are many combinations of tokamak reactor design parameters which are physically disallowed because they violate the Greenwald density limit, or the plasma beta limit, or because they are known to give rise to edge-localized modes, kink instabilities, or other MHD instabilities\cite{freidberg_designing_2015}. Likewise, many combinations of neutron star equation of state parameters are disallowed by causality considerations, nuclear saturation properties, or calculations in chiral effective field theory\cite{hebeler_equation_2013}. Other parameter combinations may be physically valid, but give rise to numerical instabilities or inaccuracies due to peculiar details of the simulation methodology. Therefore, unless one is extremely careful, applying classical dimensionality reduction techniques may very easily result in the parameter space being restricted to a substantively (or even wholly) invalid subspace, thus rendering any subsequent parameter scans scientifically meaningless.

\vspace{-0.75cm}

\section{Research Opportunities and Innovations}

\vspace{-0.5cm}

With these challenges in mind, we consider there to be a significant opportunity to develop a \textit{new, general framework} for constructing dimensionality reduction techniques that \textit{provably} enact restrictions to physically and mathematically valid parameter subspaces only, and which can be adapted to the needs of particular physical problems. Recently, we have pioneered the application of formal verification and automated theorem-proving techniques to numerical algorithms, allowing one to construct formal correctness proofs for computational physics simulations, verifying properties such as ${L^2}$ stability, flux conservation, and thermodynamic consistency of the underlying numerical scheme\cite{gorard_2025}. We advocate extending these new formal methods to automate the process of ``tagging'' regions of parameter space corresponding to physically or mathematically disallowed combinations of parameters (such as those violating fundamental density limits in plasma physics), thus enabling the development of groundbreaking new dimensionality reduction algorithms that certifiably avoid or minimize projections onto such subspaces. Such subspaces can also be optimized (through appropriate biasing of the Monte Carlo sample) to reflect the various kinds of uncertainties involved: engineering tolerances, shot-to-shot variability, experimental noise, and uncertainties in the underlying simulation model. Such a hybrid approach potentially permits us to leverage the significant performance benefits associated with modern dimensionality reduction techniques, whilst nevertheless preserving formal correctness and fidelity of the underlying physics models.

\vspace{-0.75cm}

\bibliography{inverseproblemsrefs}

\end{document}